\pdfoutput=1

\documentclass[11pt]{article}

\usepackage{acl}

\usepackage{times}
\usepackage{latexsym}

\usepackage[T1]{fontenc}

\usepackage[utf8]{inputenc}

\usepackage{microtype}

\usepackage{inconsolata}

\usepackage{graphicx}

\newcommand{\method}{\textsc{SurveyAgent}\xspace}
\newcommand{\dq}[1]{``#1''}

\usepackage{colortbl}
\usepackage{booktabs}
\usepackage{enumitem}
\usepackage{seqsplit} 
\usepackage{makecell}
\usepackage{xspace}

\definecolor{mygray}{RGB}{226, 226, 226}
\definecolor{myred}{RGB}{252, 142, 142}
\definecolor{mygreen}{RGB}{147, 255, 143}
\definecolor{myblue}{RGB}{144, 155, 255}
\definecolor{myyellow}{RGB}{253, 253, 143}
\definecolor{mypurple}{RGB}{255, 142, 250}

\title{\method: A Conversational System for Personalized and Efficient Research Survey }

\author{
    Xintao Wang\thanks{Equal Contribution.},
    Jiangjie Chen\footnotemark[1],
    Nianqi Li,
    Lida Chen,
    Xinfeng Yuan,\\
    \bf Wei Shi,
    Xuyang Ge,
    Rui Xu,
    Yanghua Xiao\\
    Fudan University\\
    \texttt{xtwang21@m.fudan.edu.cn, jjchen19@fudan.edu.cn}
}

\begin{document}
\maketitle
\begin{abstract}

In the rapidly advancing research fields such as AI, managing and staying abreast of the latest scientific literature has become a significant challenge for researchers. 
Although previous efforts have leveraged AI to assist with literature searches, paper recommendations, and question-answering, a comprehensive support system that addresses the holistic needs of researchers has been lacking. 
This paper introduces \method, a novel conversational system designed to provide personalized and efficient research survey assistance to researchers. 
\method integrates three key modules: Knowledge Management for organizing papers, Recommendation for discovering relevant literature, and Query Answering for engaging with content on a deeper level. 
This system stands out by offering a unified platform that supports researchers through various stages of their literature review process, facilitated by a conversational interface that prioritizes user interaction and personalization. 
Our evaluation demonstrates \method's effectiveness in streamlining research activities, showcasing its capability to facilitate how researchers interact with scientific literature.\footnote{Access at: \url{https://survey-agent.github.io}.}
\end{abstract}

\begin{figure}[t]
    \centering
    \includegraphics[width=\linewidth]{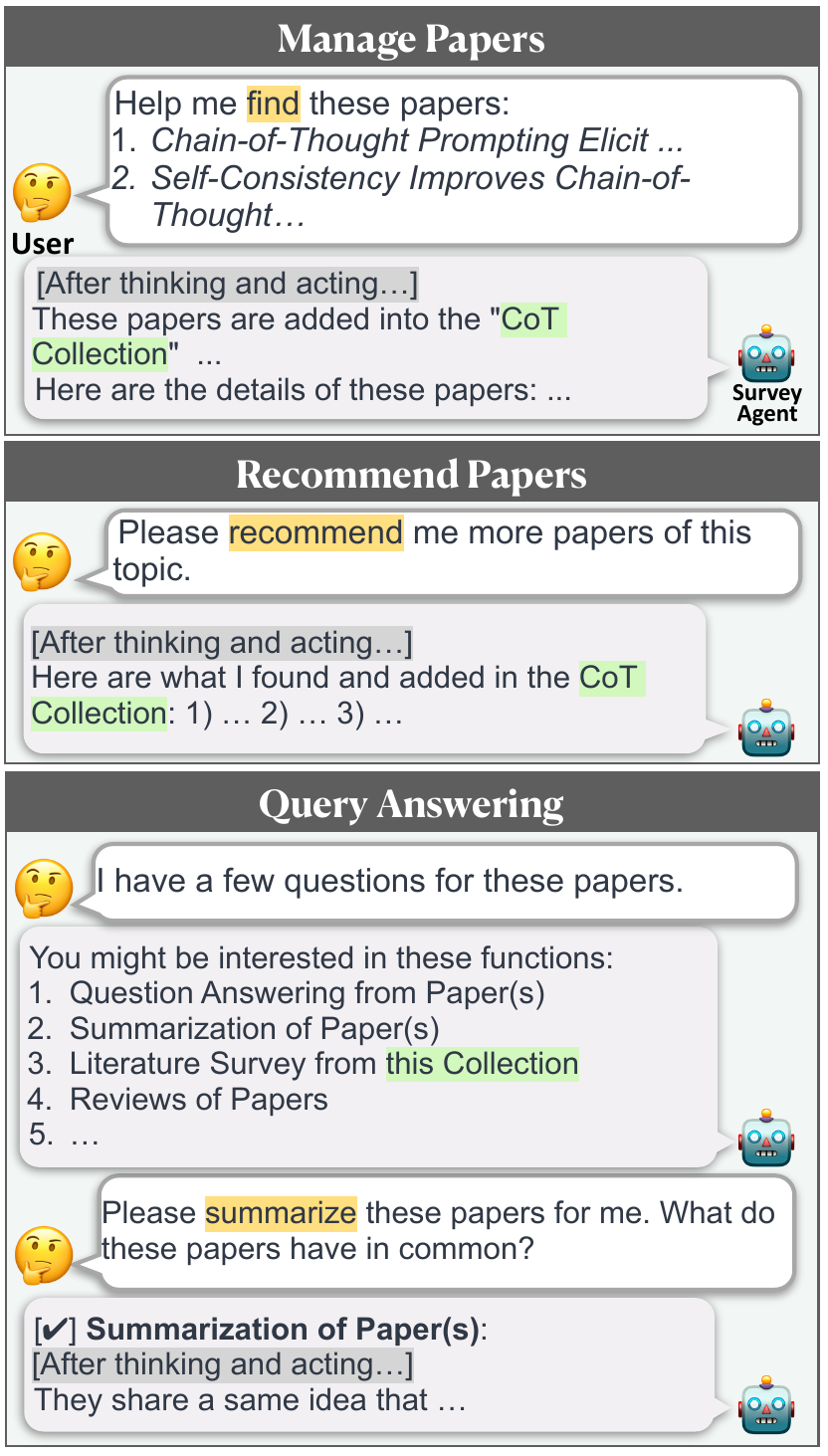}
    \caption{
    Typical use cases of \method in scientific research scenarios, where users interact with the agent through conversations.
    }
    \label{fig:1}
\end{figure}

\section{Introduction}



In fast-paced domains like Artificial Intelligence (AI), the sheer volume of scientific publications presents a formidable challenge for researchers. 
The exponential growth in literature makes it increasingly difficult to stay current with the latest advancements. 
Prior initiatives have aimed to leverage AI to alleviate this challenge, focusing on enhancing literature search efficiency through platforms like arXiv Sanity Preserver~\citep{arxiv-sanity}, improving paper recommendation systems~\citep{DBLP:journals/scientometrics/JiangMZZYD23, DBLP:journals/scientometrics/AliQKKKM22}, and developing question answering frameworks for academic content~\citep{wei2023academicgpt, DBLP:journals/corr/abs-2312-07559, kexuefm-9907}. 
These efforts are of significant use in assisting researchers with navigating the vast ocean of publications.



However, these solutions often target specific aspects of the research process without offering a holistic support system. 
This approach does not fully address the needs of researchers for a comprehensive assistant that can provide personalized, efficient navigation through the literature. 
Furthermore, despite the promising capabilities of conversational AI technologies such as ChatGPT~\citep{chatgpt,openai2023gpt4}, their application as personal research assistants remains relatively unexplored. 

In this paper, we introduce \method, a conversational system for personalized and efficient research survey to facilitate literature reviews for researchers.\footnote{In this paper, by \dq{survey} we mean the process of researchers acquiring knowledge about a research field, instead of written survey papers.}
As shown in Figure~\ref{fig:1}, \method is equipped with three modules: \textit{1) Knowledge Management module, 2) Recommendation Module and 3) Query Answering Module.} 
The {Knowledge Management Module} allows \method to find papers and organize them into collections based on the research interests of users. 
The {Recommendation Module} enables \method to search for papers through keyword-based queries and recommend similar papers, powered by the arXiv Sanity project and LLMs. 
The {Query Answering Module} empowers \method to assist users with various queries about specified papers, such as question answering, summarization and providing reviews.  
Hence, \method provides comprehensive academic support for researchers. 
With both quantitative experiments and qualitative case studies, we validate the effectiveness of \method in action planning, paper recommendation, and query answering. 

Our contributions can be summarized as follows:
\begin{enumerate}[noitemsep]
    \item We propose a general framework of research-assistance agents empowered with knowledge management, recommendation, and query-answering capabilities. 
    
    \item We design \method, a first-of-its-kind system for research surveys with a conversational interface and a comprehensive corpus of academic papers. 
    Our demo, code, and data are publicly available.

    \item Our experiments validate the effectiveness of \method in assisting research survey. 

\end{enumerate}

\section{Related Work}

\subsection{Language Agents}
Recently, LLMs have been rapidly developing towards artificial general intelligence, with lots of emergent abilities unlocked, including instruction following~\citep{NEURIPS2022_b1efde53}, in-context learning~\citep{brown2020language}, chain-of-thought reasoning~\citep{wei2022chain}.
This progress has significantly boosted the development of language agents. 
Based on LLMs, language agents empower vanilla LLMs with external modules for memory~\citep{Park2023GenerativeAgents}, planning~\citep{yao2023tree}, and action (or tool-using)~\citep{yao2023react}.
Hence, language agents have been applied to autonomously solve various problems, including question answering~\citep{yao2023react}, playing games~\citep{chen2023put}, and processing complicated requests such as placing orders or booking tickets~\citep{xie2024travelplanner}.

\subsection{Paper Querying}
Numerous past studies have explored AI systems for handling queries through scholarly articles, which can be roughly divided into \textit{paper-specific methods} and \textit{retrieval-augmented methods}. 
Paper-specific methods involve answering questions within the context of a paper provided by the user.
AcademicGPT~\citep{wei2023academicgpt} is a domain-specific LLM proposed to enhance academic capabilities, with applications developed for question answering, paper reading, review, and content generation.
PaperQA~\citep{DBLP:journals/corr/abs-2312-07559} is a retrieval-augmented agent that dynamically employs modular tools for searching papers, gathering evidence, and generating answers by leveraging LLMs to address scientific questions based on the latest literature.
Recently, Cool Papers~\citep{kexuefm-9907} leverages LLMs to provide overviews of daily arXiv papers for efficient paper skimming.
Retrieval-augmented methods, however, do not assume user-provided papers and include a paper retrieval step instead, such as ChatPaper~\footnote{https://chatwithpaper.org/}, AMiner Chat~\footnote{https://www.aminer.cn/}, and scite.ai~\footnote{https://www.scite.ai}.

\subsection{Paper Search and Recommendation}
Recommending relevant papers to researchers has long been a widely focused problem to facilitate scientific progress. 
TAPRec~\citep{DBLP:journals/scientometrics/JiangMZZYD23} revolves around addressing two core issues in paper recommendation systems: dynamic interest modeling and the cold start problem, proposing a novel time-aware recommendation model based on attention mechanisms and temporal convolutional networks.
SPR-SMN~\citep{DBLP:journals/scientometrics/AliQKKKM22} is a novel scientific paper recommendation model that effectively captures researchers' preferences and long-range dependencies, outperforming competing models in tests on real-life datasets.
arXiv Sanity~\citep{arxiv-sanity} adopts support vector machine (SVM) over tf-idf features to recommend papers similar to user-tagged papers.
Recently, gpt-paper-assistant~\footnote{https://github.com/tatsu-lab/gpt\_paper\_assistant} prompts LLMs to filter daily arXiv papers according to user interest prompts. 

Our recommendation module is based on arXiv Sanity for the term-based recommendation, followed by semantic filtering via LLMs.

\section{Methodology}
In this section, we present \method. 
We start with the agent framework (Sec~\ref{sec:framework}) and modules (Sec~\ref{sec:actions}), followed by the paper corpus (Sec~\ref{sec:papers}) and user interface (Sec~\ref{sec:ui}).

\begin{figure*}[t]
    \centering
    \includegraphics[width=\linewidth]{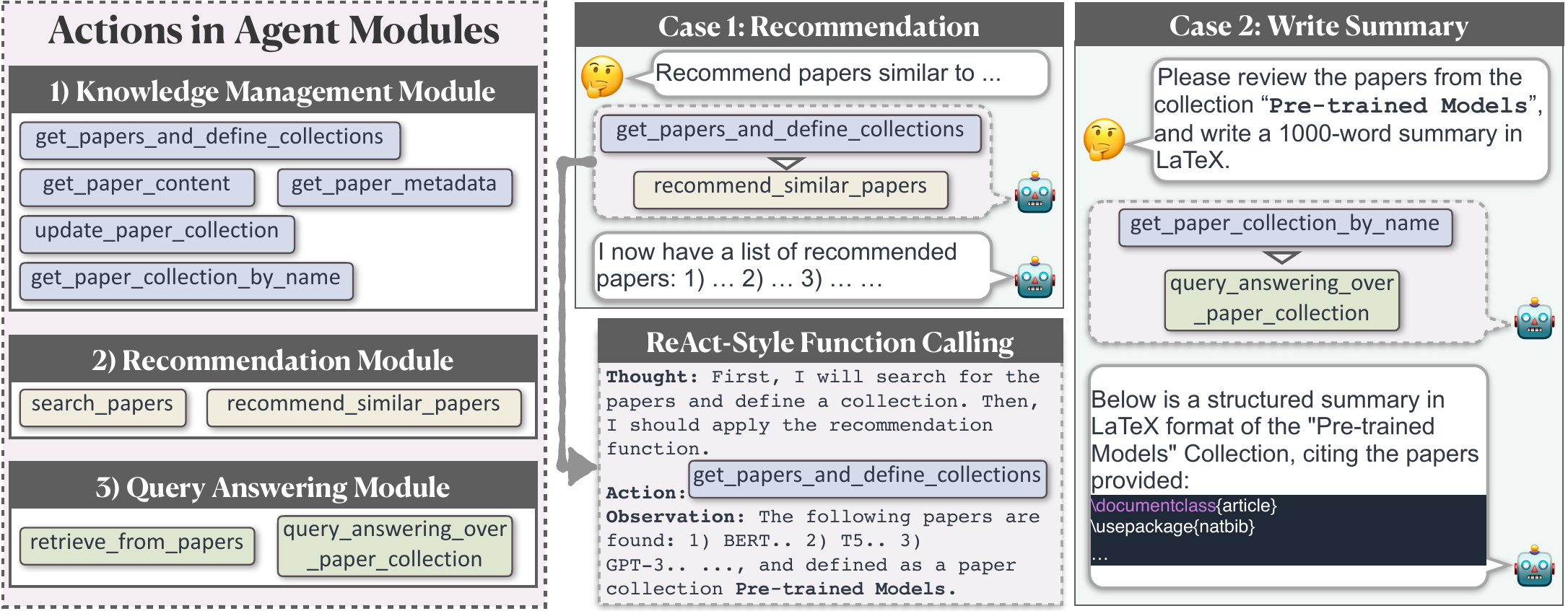}
    \caption{An overview of \method. It adopts the ReAct framework, with key modules and actions displayed on the left. We demonstrate the autonomous workflow of \method in assisting researchers with paper recommendation (Case 1) and literature summarization (Case 2).  }
    \label{fig:main}
\end{figure*}

\subsection{Agent Framework}
\label{sec:framework}

\method is an autonomous language agent based on the ReAct framework~\citep{yao2023react}, as shown in Figure ~\ref{fig:main}. 
This framework alternates between reasoning and action steps until it satisfies a user's request. 
For example, if a user requests, \dq{Please recommend some papers similar to <provided papers>.}, \method will autonomously plan and perform a series of actions, starting with searching the provided papers in the database and then making recommendations.
We equip \method with modules supporting various actions for research assistance, which are detailed in Sec~\ref{sec:actions}. 

Our implementation of \method is based on LangChain, a general framework for developing language agents.\footnote{https://github.com/langchain-ai/langchain}
The detailed prompts are listed in Appendix~\ref{sec:prompts}.

\subsection{Modules and Actions} 
\label{sec:actions}

\method offers comprehensive research assistance through three key modules: the knowledge management module, recommendation module, and query answering module, as is depicted in Fig ~\ref{fig:main}.  
Each module includes several actions. 
The detailed schemas and descriptions of our actions are listed in Table ~\ref{tab:prompts} in the Appendix.

\paragraph{Knowledge Management Module} 
The basic tasks of \method are paper management for users.  
In \method, we define the \textit{paper collection} variables, namely sets of papers, which serve as the input and output of many actions. 
Actions in this module are designed for searching for papers and paper collections within the database, as well as managing the collections. 
The detailed actions are: 
\begin{enumerate}[noitemsep]
    \item \texttt{get\_papers\_and\_define\_collections}: 
    This function takes a list of paper titles, matches them within the database, and defines them as a collection. 

    \item \texttt{get\_paper\_content}: This function retrieves the content of a paper, either abstracts, introductions, or full texts.

    \item \texttt{get\_paper\_metadata}: This function retrieves the metadata of a paper, encompassing its title, authors, publication year, and URL.

    \item \texttt{update\_paper\_collection}:
    This function adds or removes selected papers in a source collection into or from the target collection.

    \item \texttt{get\_paper\_collection\_by\_name}: 
    This function searches for a paper collection and displays information its the papers. 

\end{enumerate}

\paragraph{Recommendation Module}
Paper searching and recommendations for individualized research interests are crucial for researchers. 
Hence, we equip \method with the recommendation module, supporting the following actions:

\begin{enumerate}[noitemsep]
    \item \texttt{search\_papers}: 
    This function searches for papers based on keyword-based queries. 
    \item \texttt{recommend\_similar\_papers}: This function recommends papers similar to papers in a user-specified collection. 
\end{enumerate}


We implement these actions based on arXiv Sanity~\citep{arxiv-sanity}~\footnote{https://github.com/karpathy/arxiv-sanity-lite}, an open-source system widely recognized for its effectiveness and efficiency in searching for and recommending arXiv papers.
It applies SVMs over tf-idf features and enables paper recommendations through user-defined \textit{tags}, akin to our concept of paper collections. 
Users tag interested papers with topics like \dq{language agents} or \dq{reinforcement learning}, and arXiv Sanity recommends relevant papers similar to those with a specific tag. 
Besides, it offers the option to filter results by publication date.

However, as arXiv Sanity relies on word overlap, its results may be semantically unrelated. 
For example, searching for \dq{character role-playing of LLMs} might yield related papers on \dq{character (word) detection in computer vision} as well. 
To refine these results, we incorporate LLMs for an additional layer of filtering. 
Following gpt-paper-assistant~\citep{gpt-paper-assistant}, 
we prompt LLMs to score the relevance between the suggested papers and the search intent. 
Then, we rerank the papers based on the relevance score and keep only the most relevant papers.

\paragraph{Query Answering Module}
As a personalized research assistant, the most important feature of \method is to support users by solving their various queries about papers, such as question answering or academic writing. 
This module hence provides the following actions: 

\begin{enumerate}[noitemsep]
    \item \texttt{retrieve\_from\_papers}: 
    This function retrieves information relevant to user queries from the paper corpus based on BM25, particularly for questions about specific statements in papers.
    \item \texttt{query\_answering\_over\_paper\_collection}: This function aims at solving various user queries on papers from a specified collection, such as question answering or summarization.     
\end{enumerate}


For \texttt{query\_based\_on\_paper\_collection}, we face the challenge of super-long context in academic papers, where a research paper typically contains more than 10,000 tokens. 
Recent advancements in LLMs have made significant improvements in the context length of LLMs, such as GPT-4~\citep{openai2023gpt4}, Gemini and Mistral\cite{jiang2023mistral}.
Leveraging these long-context LLMs, we are able to input the entire text of several papers directly into the context window.
Nevertheless, the context length of current LLMs, especially the open-source models, is still insufficient for handling more papers. 
Moreover, their ability to understand and reason over scientific materials in long contexts remains to be improved.

Therefore, we propose a chunk-based approach for \texttt{query\_based\_on\_paper\_collection}. 
When the input exceeds the context limit of LLMs, we divide the papers into multiple segments using the Langchain splitter\cite{langchain}, controlling the size of each segment adhering to the context limit.
Subsequently, before responding to the question, we prompt LLMs to identify and filter out irrelevant segments based on the user query. 
Then, we gather responses from the remaining segments, with each segment potentially providing a partial answer to the query.
Finally, we merge these segment-level responses to form a global answer.

\subsection{Paper Processing}
\label{sec:papers}

\paragraph{Paper Sources}
In this paper, we mainly focus on research in the field of natural language processing (NLP). 
Hence, we collect papers published at ICML, ICLR, NeurIPS, and the ACL Anthology, as well as arXiv papers with cs:CL (Computation and Language) subjects. 
We keep only papers published since 2018, considering the rapid development of NLP research. 
The paper dataset is automatically updated every day to include newly published papers from arXiv.

\paragraph{Paper Schema} 
In the database~\footnote{We adopt Elasticsearch for database implementation.}, the papers are represented with the following schema: title, abstract, authors, institution, source, published date, year, URL, introduction, conclusion, and full text.

\paragraph{Paper Parsing}
The papers in our database are parsed by SciPDF\footnote{https://github.com/titipata/scipdf\_parser} parser, a Python parser for scientific PDF based on GROBID.
It provides detailed and well-structured parsing results, including the title, abstract, structured sections of content, references, and visual figures.
The structured sections are merged to obtain the full texts. 
To reduce unrelated information, we have excluded the references and any subsequent sections in the full text.

\subsection{User Interface}
\label{sec:ui}

We develop an interactive web user interface for \method based on HuggingChat\footnote{https://github.com/huggingface/chat-ui}, featuring the SvelteKit\footnote{https://kit.svelte.dev/} server-side rendering framework. The web interface aims to provide a user-friendly approach to interact with \method, with colorful and well-structured visualization.
This interface is mainly composed of a chat area, a session management system, and a user management system. 
Users can perform multi-round chat with markdown syntax around the queried papers, and the chat histories will be automatically recorded. Sessions of different users are separated for the sake of user privacy.

\section{Experiments}
In this section, we present the results of our experiments to give a quantitative analysis of two core capabilities of \method, namely action planning and paper recommendation. 
For qualitative case analysis, please refer to Appendix~\ref{sec:case_study}

\begin{table*}[t]\centering
    \begin{tabular}{lcccc}
        \toprule
        \textbf{Action} &\textbf{Recall (\%)}& \textbf{Precision (\%)} & \textbf{F1-score (\%)} & \textbf{Param (\%)}  \\
        \midrule
        \texttt{get\_papers\_and\_define\_collections} & 100.00 & 100.00 & 100.00 & 100.00 \\
        \texttt{get\_papercollection\_by\_name} & 100.00 & 89.28 & 94.33 & 100.00 \\
        \texttt{get\_paper\_content} & 100.00 & 100.00 & 100.00 & 100.00 \\
        \texttt{get\_paper\_metadata} & 100.00 & 100.00 & 100.00 & 100.00 \\
        \texttt{update\_paper\_collection} & 88.00 & 100.00 & 93.61 & 96.59 \\
        \texttt{retrieve\_from\_papers} & 40.00 & 100.00 & 57.14 & 100.00 \\
        \texttt{search\_papers} & 100.00 & 89.28 & 94.33 & 100.00 \\
        \texttt{recommend\_similar\_papers} & 96.00 & 100.00 & 97.95 & 100.00 \\
        \texttt{query\_based\_on\_paper\_collection} & 80.00 & 100.00 & 8888 & 100.00 \\
        \midrule
        Total & 89.33 & 97.61 & 93.29 & 99.63 \\
        \bottomrule
    \end{tabular}
    \caption{Results of single action planning by \method. Generally, \method exhibits strong capabilities in accurately generating the actions and parameters.}
    \label{single-action-planning-result}
\end{table*}
\begin{table}[t]\centering
    \resizebox{\columnwidth}{!}{
    \begin{tabular}{lcccc}
        \toprule
        ~ & \textbf{Commit} & \textbf{3 Seeds} & \textbf{5 Seeds} &  \textbf{9 Seeds}\\
        \midrule
        {arXiv Sanity} & 0.174 & 0.133 & 0.152 & 0.194 \\
        \small{\quad+LLM direct} & 0.184 & 0.149 & 0.176 & 0.199 \\
        \method & \textbf{0.205} & \textbf{0.171} & \textbf{0.184} & \textbf{0.205} \\
        \bottomrule
    \end{tabular}
    }
    \caption{Paper recommendation performance (HR@10) of \method. \dq{Commit} means using first-time committed papers as the seed to start recommendation. \dq{K Seeds} denotes the number of seed papers.}
    \label{recommendation-result}
\end{table}

\subsection{Action Planning}

\paragraph{Single Action Planning}
First, we evaluate \method's ability in understanding and applying individual actions. 
To this end, we construct the single action dataset using GPT-4.
This dataset contains 25 test samples for each action (225 samples in total), where each sample consists of a query, an action, and the corresponding input parameters.

We conduct experiments on the dataset. 
We report the precision, recall, and F1 score for action selection, and accuracy of the input parameters.
The results are shown in Table~\ref{single-action-planning-result}.
Overall, our system is capable of accurately generating the intended actions and parameters, achieving the averaged F1 score of 93\% and parameter accuracy of 99\%. 
We observe that several queries are misclassified as \texttt{search\_papers} while the labels are \texttt{retrieve\_from\_papers}, 
which results in a poor recall of \texttt{retrieve\_from\_papers}.
This is because the intended \dq{statements} for \texttt{retrieve\_from\_papers} are also good queries for \texttt{search\_papers}.


\paragraph{Multiple Action Planning}
Furthermore, we evaluate the multiple action planning ability of \method. 
The multiple action dataset is constructed with the following steps.
First, we construct 150 samples of multi-action trajectories using GPT-4.
Then, we manually check the reasonableness and plausibility of the query, and finally retain 50 testing queries, where each sample contains a query and its corresponding series of actions.
The samples are illustrated in Appendix~\ref{sec:case_study}.

We report three metrics for this task, including the single-action accuracy, full-trajectory accuracy and the edit distance compared with the ground truth, which are 85.83\%, 52.00\% and 0.42 for \method respectively.  
However, considering the variability of multiple action planning, the labels are unnecessarily the only answers.
Therefore, we further conduct human evaluation and observe an 82\% full-trajectory accuracy, largely surpassing the 52.00\% result as measured by ground truth labels. 
There are mainly two reasons behind the improvement: 
First, the system choose to use \texttt{get\_papercollection\_by\_name} instead of a portion of the \texttt{search\_papers} operation, which saves retrieval consumption;
Second, the system may define \texttt{get\_papers\_and\_define\_collections} on its own after \texttt{search\_paper} or \texttt{retrieve\_from\_papers}, which facilitates subsequent re-access to this part of the paper.
Overall, \method is highly capable of generating multi-action plans for academic queries.

\subsection{Paper Recommendation}

\paragraph{Dataset Construction}
We evaluate \method's ability to recommend relevant papers based on user-provided papers.
Prior datasets for paper recommendations fall short of our evaluation needs~\cite{DBLP:conf/aaai/GuoCZLDH20,DBLP:conf/ijcai/WangCL13,DBLP:conf/jcdl/SugiyamaK13}.
Hence, we construct a new dataset, based on \dq{awesome-x-papers} repositories on Github that gather papers for specific domains.
Concretely, we choose repositories that include ``Awesome LLM'' in their names, and categorize the specific subjects based on the subheadings found within each repository.
We construct the initial collection from the papers included in the first commit where the subject has more than three papers, and the paper collection from the most recent commit serves as our target collection.
After filtering out data where the difference between the target collection and the initial collection is less than five papers, we ultimately obtain 30 samples.
Also, We use the paper collection from the final commit as the target collection, construct initial collections by randomly selecting 3, 5, and 9 papers respectively, and after the filtering process, we obtain 50 samples.

\paragraph{Baselines} \method recommend papers based on arXiv Sanity, and filter the results with relevance scored by LLMs. 
For ablation study, we show the recommendation results of arXiv Sanity only, and arXiv Sanity (+LLM direct) that directly filter papers without relevance scores.


\paragraph{Recommendation Result}

To evaluate the recommendation capacity of our system, we adopt the widely used metric, top-K Hit Ratio (HR), and set the k as 10.
We present the performance of two distinct recommendation strategies on our test data, with the arXiv Sanity approach serving as our baseline.
As reported in Table~\ref{recommendation-result}, both the LLM\_direct and LLM\_score methods outperform the arXiv Sanity method, with the LLM scoring method for each paper's relevance achieving the best performance on all test data.
As the size of the initial collection increases, all methods are capable of delivering more precise paper recommendations.
It's worth noting that our recommendation strategies maintain commendable performance even when the quantity of papers in the initial collection is relatively limited.
\section{Conclusion}

In our work, we presented \method, a conversational system to facilitate how researchers engage with the vast volume of literature in research fields like AI.
Through our experiments, we show that \method exhibited exceptional capability in understanding and executing both single and multi-step action plans with high accuracy, while showcasing an advanced capability in providing precise paper recommendations, outstripping traditional methods like arXiv Sanity. 
Notably, its nuanced LLM-based scoring for paper relevance proved its effectiveness across different sizes of initial collections.
These results not only highlight \method's potential to significantly enhance research efficiency by streamlining the literature review process but also mark a significant step forward in the application of conversational AI for academic research assistance, promising a future where researchers can navigate the ever-expanding sea of publications with ease and precision.



\bibliography{reference}

\clearpage
\appendix

\section{Case Study}\label{sec:case_study}

\begin{table*}[t]\centering
    \resizebox{0.98\textwidth}{!}{
    \large
    \begin{tabular}{l|l}
        \toprule
        \toprule
        \multicolumn{1}{c|}{Query} & \multicolumn{1}{c}{Answer} \\
        \midrule
        \multicolumn{2}{c}{Single Action}\\
        \midrule
        \makecell[l]{I have the following three documents: 1) MAmmoTH,\\ 2) ToRA. Save the above documents as a group named \\``Mathematical Reasoning''} &                     \makecell[l]{Action: \textit{get\_papers\_and\_define\_collections}\\The documents have been saved as a group named ``Mathematical Reasoning''.}\\
        \midrule
        \makecell[l]{What papers are included in ``Mathematical Reasoning''} &                                      \makecell[l]{Action: \textit{get\_papercollection\_by\_name}\\The ``Mathematical Reasoning'' collection includes the following papers: \\1. "MAmmoTH" by Xiang Yue, Xingwei Qu... in 2023. \\2. "ToRA" by Ke Wang, Houxing Ren... published in 2023.}\\
        \midrule
        \makecell[l]{Find the abstract of ``MAmmoTH Building Math \\Generalist Models through Hybrid Instruction Tuning''} &     \makecell[l]{Action: \textit{get\_paper\_content}\\The abstract of "MAmmoTH: Building Math Generalist Models through \\Hybrid Instruction Tuning" introduces MAmmoTH, a series of open-source \\large language models designed for general math...}\\
        \midrule
        \makecell[l]{Find the authors of ``MAmmoTH: Building Math \\Generalist Models through Hybrid Instruction Tuning''} & \makecell[l]{Action: \textit{get\_paper\_metadata}\\The authors of the paper "MAmmoTH: Building Math Generalist Models \\through Hybrid Instruction Tuning" are Xiang Yue...}\\
        \midrule
        \makecell[l]{Add the first paper in ``chatbot\_personality\_collection'' \\to the ``Mathematical Reasoning'' group} &                        
        \makecell[l]{Action: \textit{update\_paper\_collection}\\The first paper has been successfully added.}\\
        \midrule
        \makecell[l]{Search for papers that contain the statement: ``Artificial \\Intelligence can enhance healthcare delivery''} & \makecell[l]{Action: \textit{retrieve\_from\_papers}\\The following papers contain discussions relevant to the statement ``Artificial \\Intelligence can significantly enhance healthcare delivery'': 1. Using Large \\Pre-Trained Language Model to Assist FDA in Premarket Medical Device by \\Zongzhe Xu (2022) discusses the use of AI in healthcare for more accurate \\diagnoses and ...}\\
        \midrule
        \makecell[l]{Find a survey of mathematical reasoning} & \makecell[l]{Action: \textit{search\_papers}\\I have found several papers related to a survey of mathematical reasoning. \\Some of the notable papers include: `A Survey of Deep Learning for \\Mathematical Reasoning by Pan Lu et al., 2023...}\\
        \midrule
        \makecell[l]{Recommend some papers similar with collection \\``Mathematical Reasoning''} & \makecell[l]{Action: \textit{recommend\_similar\_papers}\\Here are some papers similar to those in the ``Mathematical Reasoning'': \\``Natural Language Reasoning, A Survey'' by Fei Yu et al., 2023.``A Survey \\in Mathematical Language Processing'' by Jordan Meadows et al., 2022...}\\
        \midrule
        \makecell[l]{Based on the papers in ``Mathematical Reasoning'',\\ what are the current methods that can achieve zero-shot \\math reasoning?} & \makecell[l]{Action: \textit{query\_based\_on\_paper\_collection}\\The current methods that can achieve zero-shot math reasoning, as \\described in the papers from the "Mathematical Reasoning" collection, \\include: 1. Chain-of-Thought Prompting (CoT): This method encourages \\models to solve problems incrementally on a scratchpad...}\\
        \midrule
        \multicolumn{2}{c}{Multi Action}\\
        \midrule
        \makecell[l]{I am investigating the research field of role-playing AI. \\There are several key papers in this field: 1. ChatHaruhi \\2. RoleLLM 3. Character-LLM 4. CharacterGLM. \\Please recommend me relevant papers in this field.} & \makecell[l]{Action: \textit{get\_papers\_and\_define\_collections}, \textit{recommend\_similar\_papers}\\Here are some papers that are relevant to the field of role-playing AI: \\1. Character Set Construction for Chinese Language Learning by Chak Yan \\Yeung, John Lee (2021) 2. The Evolution of Popularity and Images of \\Characters in Marvel Cinematic Universe Fanfictions by Fan Bu (2018)...}\\
        \midrule
        \makecell[l]{I would like to know more about the machine learning \\in healthcare. Could you find some papers related to \\this subject and then summarize their findings?} & \makecell[l]{Action: \textit{search\_papers}, \textit{query\_based\_on\_paper\_collection}\\The main findings from the collection of papers on machine learning \\in healthcare can be summarized as follows: 1. Information Extraction \\and NLP in Scientific Domains:   - Machine learning pipelines are \\effective for structured information extraction from scientific literature...}\\
        \bottomrule
        \bottomrule
    \end{tabular}
    }
    \caption{The examples for the correct action planning.}
    \label{case-study-correct-examples}
\end{table*}

Table~\ref{case-study-correct-examples} shows the samples used to test single-action and multi-action as well as the system's responses.
The results are the same as those we presented in the experimental section, that is, our system understands the user's intention well and takes the correct action.
However, for a very small number of samples, our system still has errors.
And we show some typical samples of errors in Table~\ref{case-study-error-examples}.
For single action, the model's errors mainly come from the confusion between \texttt{retrieve\_from\_papers} and \texttt{search\_papers}, because some questions are difficult to distinguish whether they are ``specific statements'' or ``keywords''.
At the same time, due to the long text and the large collection of papers, it is undeniable that some samples will take too much time during the test, which will also lead to failure.
For multi-action, in addition to the problems mentioned above, there are also a few action planning errors, which are caused by the model misinterpreting the user's input.

\begin{table*}[t]\centering
    \resizebox{0.98\textwidth}{!}{
    \large
    \begin{tabular}{l|l}
        \toprule
        \toprule
        \multicolumn{1}{c|}{Query} & \multicolumn{1}{c}{Answer} \\
        \midrule
        \makecell[l]{Retrieve papers that discuss the concept: Augmented \\Reality can improve learning outcomes in education} &  \makecell[l]{Action: \textit{search\_papers}\\Error: When asking for a specific statement, $retrieve\_papers$ should be used \\instead of $search\_papers$}\\
        \midrule
        \makecell[l]{What are the key takeaways from the Blockchain \\Technology collection in terms of potential use cases?} & \makecell[l]{Action: \textit{query\_based\_on\_paper\_collection}\\Error: No results were returned for 20 minutes due to the length of the text that \\the model needed to read}\\
        \midrule
        \makecell[l]{I'm interested in the topic Natural Language Processing.\\Search for papers on this topic, add the first three to my \\NLP Papers collection, and then show me this collection.} & \makecell[l]{Action: \textit{search\_papers}, \textit{get\_papers\_and\_define\_collections}, \\\textit{get\_papercollection\_by\_name}\\Error: Misunderstanding of the user's intent, the system constructs a new paper \\collection instead of updating the paper collection.}\\
        \bottomrule
        \bottomrule
    \end{tabular}
    }
    \caption{The examples for the error action planning.}
    \label{case-study-error-examples}
\end{table*}

\section{Prompts}
\label{sec:prompts}

We list the detailed prompts for \method in Table~\ref{tab:prompts}, including the system prompt, and tool using examples and action descriptions.

\begin{table*}[h]
\centering
\resizebox{\linewidth}{!}{
\begin{tabular}{p{3.7cm}|p{19.7cm}}
\toprule
\multicolumn{2}{c}{\textbf{Prompts}} \\
\midrule
    \rowcolor[rgb]{ .949,  .953,  .961} \multicolumn{2}{c}{\textit{Instructions}} \\
    {\textbf{System Prompt}}&
    You are Survey Agent, an AI-driven tool expertly crafted for researchers to facilitate their exploration and analysis of academic literature. With a suite of advanced functions, you excel in organizing, retrieving and recommending research papers, and answering questions based on these papers.
    
As Survey Agent, you serve as a vital assistant to  researchers, simplifying the task of navigating through the extensive and complex domain of academic literature, and delivering tailored, relevant, and accurate insights. In a nutshell, you should always answer the user's academic queries as best you can.

You shoulde use tools for paper retrieval, paper collection management, paper recommendation, and question answering. Don\'t answer it yourself if you can use a tool to answer it. Specifically, you have access to the following tools:

\{tools\}

For single parameter input, please input directly; for multiple parameter input, please use dict format to input.

Here are some simple examples to tell you when to use which tools:

\{tool\_using\_example\}

Use the following format:

Query: the input query for which you must provide a natural language answer
Thought: you should always think about what to do, step by step
Action: the action to take, should be one of [\{tool\_names\}]
Action Input: the input to the action. For boolean parameters, use lowercase (true / false). 
Observation: the result of the action
... (this Thought/Action/Action Input/Observation can repeat N times.)
Thought: I now know the final answer
Final Answer: the final answer to the original input question (do not repeat large blocks of content that is present in the Observation.)

\{chat\_history\}

Question: \{input\}
\{agent\_scratchpad\}
    \\ \midrule

    {\textbf{Tool Using Examples}} &
    Query: Find similar articles to <certain papers>.
    
Steps: Using \"get\_papers\_and\_define\_collections\" to define collection of given papers. Then, using \"recommend\_similar\_papers\" to get similar articles.

Query: Find some papers related to <a certain field>.

Steps: Using \"get\_papercollection\_by\_name\" to find if there is already a collection for this field. If not, using \"search\_papers\" to find related papers.

Query: Summarize the papers related to <field>.

Steps: Using \"get\_papercollection\_by\_name\" to find if there is already a collection for this field. If not, using \"search\_papers\" to find related papers. Then, using \"query\_based\_on\_paper\_collection\" to summarize.

Query: Which paper proves <certain conclusion>.

Steps: Using \"retrieve\_from\_papers\" find paper about the conclusion. Then, using \"query\_based\_on\_paper\_collection\" find answer.

Query: Please summarize this <paper collection> and write a summary.

Steps: Using \"query\_based\_on\_paper\_collection\" to answer.
    \\ \midrule

    \rowcolor[rgb]{ .949,  .953,  .961} \multicolumn{2}{c}{\textit{Action Descriptions}} \\
    \textbf{\seqsplit{get\_papers\_and\_define\_collections}} &
    get\_papers\_and\_define\_collections(paper\_titles: List[str], paper\_collection\_name: str) -> str - This function processes a list of paper titles, matches them with corresponding entries in the database, and defines a collection of papers under a specified name.
        Note that:
            1. If certain papers are not found, do not attempt to use the search\_papers function again to look for those papers.
            2. Only use this function when the user inputs a list of paper titles. Do not use it without explicit intention from the user. 
    \\ \midrule

    \textbf{\seqsplit{get\_paper\_content}}&
    
    get\_paper\_content(paper\_name: str, mode: str) -> str - Retrieve the content of a paper. Set 'mode' as 'full' for the full paper, or 'abstract' for the abstract.
    \\ \midrule

    \textbf{\seqsplit{get\_paper\_metadata}} &
    get\_paper\_metadata(paper\_name: str) -> str - Retrieve the metadata of a paper, including its title, authors, year and url. 
    \\ \midrule

    \textbf{\seqsplit{get\_papercollection\_by\_name}}&
    get\_papercollection\_by\_name(collection\_name: str) -> str - Retrieve a specified paper collection by its name, display the paper collection's name and information of its papers. Only use this function when the user explicitly asks for information about the collection. Avoid using this when the user poses a request about the collection, in which case the agent should use 'query\_based\_on\_paper\_collection' instead. 
    \\ \midrule

    \textbf{\seqsplit{update\_paper\_collection}} &
    update\_paper\_collection(target\_collection\_name: str, source\_collection\_name: str, paper\_indexes: str, action: str) -> bool - Updates the target paper collection based on a specified action ('add' or 'del') and paper indices (Indices start from 0. The format should be comma-separated, with ranges indicated by a dash, e.g., "0, 2-4") from the source collection.
    \\ \midrule

    \textbf{\seqsplit{retrieve\_from\_papers}} &
    retrieve\_from\_papers(query: str) -> str - Retrieve the most relevant content in papers based on a given query, using the BM25 retrieval algorithm. Output the relevant paper and content. This function should be used when the query is about a specific statement, rather than being composed of keywords.
    \\ \midrule

    \textbf{\seqsplit{search\_papers}} &
     search\_papers(query: str, time\_filter: str = '') -> str - Searches for papers based on a given query. Optionally filter papers that were published 'time\_filter' days ago.
        The query should consist of keywords rather than a complete paper title. If the user's input seems like a paper title, the agent should use 'get\_papers\_and\_define\_collections'.
    \\ \midrule

    \textbf{\seqsplit{recommend\_similar\_papers}} &
    recommend\_similar\_papers(collection\_name: str, time\_filter: str = '') -> str - Recommends papers similar to those in a specified collection. Optionally filter papers that were published 'time\_filter' days ago.
        Note that:
        1. Only use this function when the user explicitly asks for recommendation.
    \\ \midrule

    \textbf{\seqsplit{query\_based\_on\_paper\_collection}} &
query\_based\_on\_paper\_collection(paper\_list\_name, query, content\_type, model\_type='large', chunk: bool = False) -> str - When the user poses a question or request concerning a specific paper collection, the agent should use this action to generate the answer. This action includes the 'get\_papercollection\_by\_name' function. Therefore, the agent should call this action directly instead of first invoking 'get\_papercollection\_by\_name'.
        Note that:
        1. 'content\_type' denotes which part of the papers would be used to answer the query. Choose from "abstract", "intro" or "full" for the abstract, introduction or the full text of the papers respectively.
        2. 'model\_type' denotes which kinds of LLMs would be used to answer the query. Use "large" by default to use Gemini-pro, or use "small" for smaller open-source models if specified by the user.
        3. 'chunk' denotes applying the 'chunk-and-merge' algorithm. Set it as False by default unless it is specified by the user.
        4. If the user-specified paper collection is not found, the agent should finish this round and wait for user instructions.   
        \\
\bottomrule

\end{tabular}}
\caption{Prompts for \method. }
\label{tab:prompts}
\end{table*}

\end{document}